\documentclass[letterpaper, 10 pt, conference]{ieeeconf}  
\usepackage[utf8]{inputenc}
\usepackage[T1]{fontenc}
\usepackage{graphicx}
\usepackage{booktabs}
\usepackage{caption}
\usepackage{amsmath}
\usepackage[table,dvipsnames]{xcolor}
\usepackage{dirtytalk}

\usepackage[normalem]{ulem}
\usepackage{xcolor}
\usepackage{footmisc}
\usepackage{cite}
\usepackage{hyperref}

\IEEEoverridecommandlockouts                              

\title{\LARGE \bf
Impact of Different Failures on a Robot's Perceived Reliability
}

\author{Andrew Violette$^{1}$, Zhanxin Wu$^{1}$, Haruki Nishimura$^{2}$, Masha Itkina$^{2}$, \\ Leticia Priebe Rocha$^{2}$, Mark Zolotas$^{2}$, Guy Hoffman$^{3}$, Hadas Kress-Gazit$^{3}$ 
\thanks{$^{1}$Department of Computer Science, Cornell University
       }%
\thanks{$^{2}$Toyota Research Institute (TRI).   }
\thanks{$^{3}$ Sibley School of Mechanical and Aerospace Engineering, Cornell University
       }
\thanks{TRI provided funds to assist the authors with their research; this article solely reflects the opinions and conclusions of its authors, and not TRI or any other Toyota entity.}%
}

\begin{document}

\maketitle
\thispagestyle{empty}
\pagestyle{empty}

\begin{abstract}

Robots fail, potentially leading to a loss in the robot's perceived reliability (PR), a measure correlated with trustworthiness. In this study we examine how various kinds of failures affect the PR of the robot differently, and how this measure recovers without explicit social repair actions by the robot. In a preregistered and controlled online video study, participants were asked to predict a robot's success in a pick-and-place task.  We examined manipulation failures (slips), freezing (lapses), and three types of incorrect picked objects or place goals (mistakes). Participants were shown one of 11 videos---one of five types of failure, one of five types of failure followed by a successful execution in the same video, or a successful execution video. This was followed by two additional successful execution videos. Participants bet money either on the robot or on a coin toss after each video. People’s betting patterns along with a qualitative analysis of their survey responses highlight that mistakes are less damaging to PR than slips or lapses, and some mistakes are even perceived as successes. We also see that successes immediately following a failure have the same effect on PR as successes without a preceding failure. Finally, we show that successful executions recover PR after a failure. 
Our findings highlight which robot failures are in higher need of repair in a human-robot interaction, and how trust could be recovered by robot successes. 
\end{abstract}


\section{Introduction}
\label{sec:Intro}

\begin{figure*}[t]
  \includegraphics[width=1\linewidth]{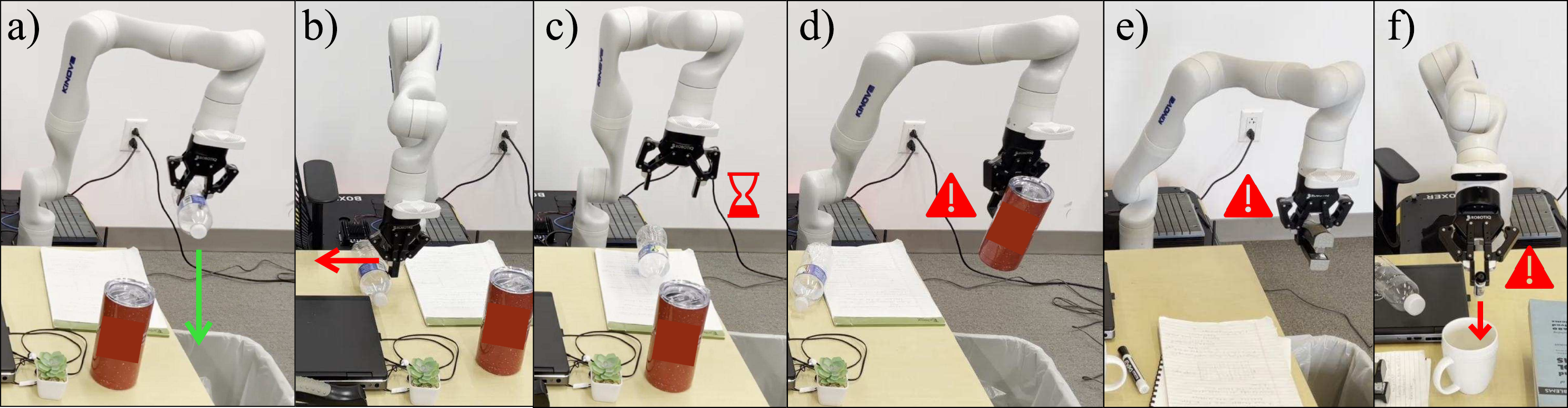}
  \caption{6 of the 11 conditions examined in this study.  From left to right: a) \textbf{Success}. The robot puts the bottle in the trash. b) \textbf{Slip}. The robot is unable to pick up the bottle. c) \textbf{Lapse}. The robot positions to pick up the bottle, then freezes for 15 seconds. d) \textbf{Mistake (Thermos)}. The robot picks up the reusable thermos and places it in the trash. e) \textbf{Mistake (Stapler)}. The robot picks up the stapler and places it into the trash. f) \textbf{Mistake (Marker)}. The robot picks up the marker and places it into the mug. Not shown: The other five conditions, which are the failure modes (b--f) immediately followed by success.}
  \label{fig:conditions}
\end{figure*}

Robots are increasingly used in the workplace, working alongside humans. Application areas include logistics~\cite{eppner2016amazonlessons},  manufacturing~\cite{arents2022smart,liu2024application}, and service industries~\cite{moriuchi2024role,gonzalez2021service}. For example, shipment warehouses have introduced robots handing objects to humans who pack orders~\cite{allgor2023algorithm}, 
and robots routinely perform deliveries in cities \cite{schneider_how_2024}. 

As these robots operate, they will invariably fail. 
Limitations to software, hardware, and the robots' interaction with the environment all lead to potential failures~\cite{steinbauer2012survey}.
This can include misidentifying an object, selecting an inappropriate action, and misinterpreting the task. For example, when instructed to grasp a red mug on the table, the robot might instead pick up a yellow mug or push the mug rather than grasp it. 

In this work, we classify robot failures based on taxonomies of human failure, in line with prior work in human-robot interaction (HRI)~\cite{kramer_human-agent_2012}. 
Reason et al. categorized human error into \textit{slips}, \textit{mistakes}, and \textit{lapses}~\cite{reason_human_1990}. Slips occur when the plan is correct, but the agent suffers a mechanical failure during execution. Mistakes occur when the plan is wrong---manipulating the wrong object, or putting the correct object in the wrong location. Lapses occur when the plan is initially correct, but is `lost' during execution---spending extended periods of time making no progress towards the goal.


Our research question is: do different types of failure have different effects on how reliable the robot seems to users? 
To answer this question, we showed users videos of different types of robot failures in a pick-and-place task. 
Each participant viewed one of 11 videos explained in Fig.~\ref{fig:conditions}---five types of Failure, five types of Failure+Success (a failure followed by a successful execution in the same video), or a successful execution video.
This experimental manipulation was followed by two successful execution videos. 

After each video, participants were asked to choose between betting on the robot’s success or on a fair (50\%) coin. If they chose the robot and the robot failed, or they lost the coin lottery, participants would lose \$2 of their study compensation. This study design allowed us to evaluate the perceived reliability (PR) of the robot using an objective measure, rather than self-report. PR is a measure correlated to trustworthiness~\cite{hancock2023how} and we evaluate it instead of trust in the context of HRI, as it can be construed without the need for anthropomorphizing the robot or introducing vulnerability.



We find that (a) mistakes are the least damaging of failures,
(b) that success immediately after failure has the same impact on PR as success without failure, and
(c) participants return to perceive the robot as reliable if it eventually succeeds, regardless of failure type.

Our findings can inform which types of failures are in higher need of repair strategies.
Slips and lapses, in particular, produce large drops in PR, suggesting that they indicate an important flaw on the robot's part. 
In contrast, fluent physical execution of a wrong plan, on a wrong object, or with a wrong task location do not seem to faze participants as much.
People also seem to be forgiving of robots that fail before eventually succeeding, even without explicit trust repair strategies.
These findings can have important implications on the design of failure recovery techniques in HRI.





\section{Related Work}
\label{sec:RelatedWork}

Taxonomies of robot failure are delineated by cause and severity \cite{honig_understanding_2018}. Cause-based taxonomies analyze the source of the failure by technical module (e.g., software, hardware) or agent (robot, human, interaction between the two) \cite{carlson_how_2005, steinbauer2012survey}. Severity definitions measure failure based on the service being provided \cite{laprie_dependability_1995} and whether the intended service can be recovered \cite{ross_demonstrating_2004, carlson_how_2005}. 
In this work, we adapt the taxonomy of human failure from \cite{reason_human_1990}, which defines errors according to their primary cognitive processing stage---mistakes occur during planning, lapses occur when the initial plan is correct but is lost during storage, and slips occur when a plan is not executed correctly.
\cite{team_careful_2025} highlights these failure modes in executions of visual language action (VLA) models: \textit{mistakes} (\say{tends to execute the wrong task}), \textit{lapses} (\say{unable to move away from its starting pose}), and \textit{slips} (\say{fails when attempting to grasp the object}).

Do different types of robot failures have different impacts on PR? The results in the literature are mixed. Slips and mistakes had the same impact on user trust for a social approach task~\cite{flook_impact_2019}, and differed only in perceived benevolence for a risk detection task \cite{kox_trust_2025}. However, different types of errors impact the efficacy of different recovery strategies~\cite{zhang_effects_2023}, and users express distinct emotional reactions depending on the severity of error~\cite{stiber_not_2020}. 
We seek to expand the understanding of different failures by comparing slips, lapses, and multiple types of mistake.

Once mistakes occur, trust recovery techniques have primarily focused on social methods---for example, by apologizing~\cite{fratczak_robot_2021}, providing a justification for the mistake~\cite{correia_exploring_2018}, ignoring the mistake \cite{engelhardt_better_2017}, or promising to do better~\cite{Karli2023}. There is also a large body of work focused on trust repair strategies for specifically slips~\cite{stiber_not_2020,zhang_effects_2023}.

Lapses and Failure+Success have only sparse representation in the literature. While delays in execution are well studied in the context of teleoperation~\cite{yang_effect_2015}, and in social interaction~\cite{shiwa_how_2009, kang_robot_2024, pelikan_managing_2023}, there is limited research into how a delay in execution impacts PR for autonomous manipulation tasks outside of a social context. As the failure type with the second-largest impact on PR, lapses require further study. There is also a gap around \textit{Failure+Success}---failure behavior followed immediately by successful task completion. This is called a physical trust repair strategy, and is most effective when performed automatically \cite{lane_robots_2024}. We examine how physical trust repair strategies impact PR for different types of failures without social intervention.





\section{Method}
\label{sec:Methods}
Our research questions are: How do different kinds of failures, specifically  slips, mistakes, and lapses, affect a robot's PR? How does this change if the robot successfully completes the task after the failure?
To investigate these questions, we pre-registered two hypotheses before running the experiment described in this paper:\footnote{The preregistration can be found at the following link: \href{https://aspredicted.org/jjgf-s7cw.pdf}{AsPredicted 233286}.
In the preregistration, we use the phrase \say{unexpected behaviors} for the construct we call here, for clarity, \say{failures}. We do not use either term in experimental materials presented to participants.}
\begin{enumerate}
    \item  When a user observes an autonomous robot completing a manipulation task, different types of failures, defined as slips, mistakes, and lapses, will each have a different impact on PR, and
 
    \item Successes will have different impacts on PR depending on whether a failure precedes the success and depending on what type of failure occurred. 
 \end{enumerate}

\begin{figure}[t]
  \centering
  \includegraphics[width=0.9\linewidth]{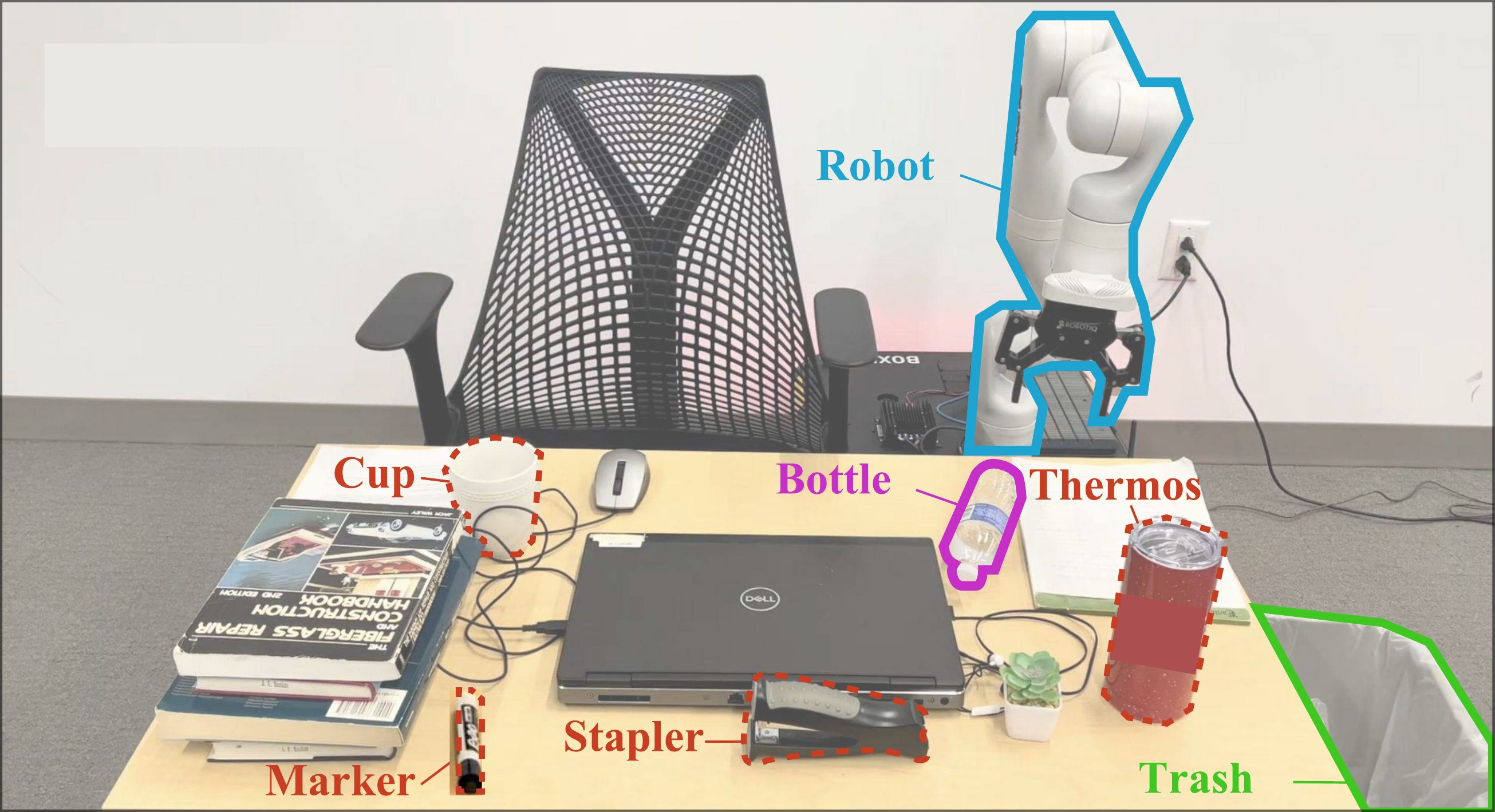}
  \caption{The experimental setup. The robot (blue) attempted to place the disposable plastic water bottle (pink) into the trash (green). In some videos, the robot interacted with other objects (dotted red). Participants saw an unlabeled version of this before placing bets. 
  }
  \label{fig:Setup}
\end{figure}

To address these research questions,  we conducted an online video study. Participants ($n$=326, 161 male, 159 female, 6 other, aged 18-78, median:37) watched videos of a robot attempting to place a disposable plastic water bottle into a trash can as shown in Fig.~\ref{fig:Setup}. All videos were of a robot arm being teleoperated to demonstrate the experiment condition, although this was not specified to the participants.

\begin{figure*}[t]
  \includegraphics[width=1\linewidth]{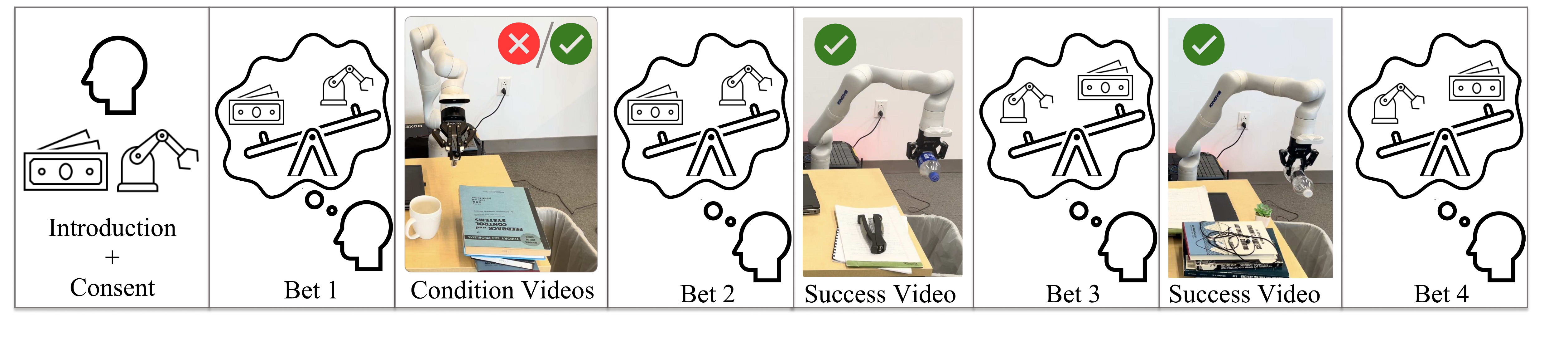}
  \caption{The flow of the experiment. Participants were asked to place bets on robot success before and after seeing three videos---one of the experimental conditions, then two success videos. 
  }
  \label{fig:ExperimentFlow}
\end{figure*}

\subsection{Experiment Flow}
\label{subsec:Flow}

The experiment flow is shown in Fig.~\ref{fig:ExperimentFlow}. First, participants gave their consent and answered a 5-point scale question on their attitude toward introducing new technology into their lives. Then, they read a description of the robot's task, the betting mechanism, and their payout.

After the introduction, participants saw an alternating sequence of betting prompts and robot execution video screens.
Before watching each video, the participant placed a bet on whether the robot would succeed, based on a description of the task and a photo of the workspace. 
They were asked whether they bet on \say{[t]he robot to put the water bottle in the trash} or \say {[a] random 50/50 coin flip}.
Each bet wagered \$2 out of their initial \$10 participant payout.

When placing their bet, they also answered a 5-point Likert item on how confident they are in their bet, from \say{not at all confident} (1) to \say{very confident} (5). In addition, they provided a short open-ended response explaining their reasoning for both the choice and their confidence level.

The experimental manipulation occurred by randomly assigning participants to see one of eleven \say{condition videos}, (third from the left in Fig.~\ref{fig:ExperimentFlow}).

After seeing the condition video, participants also evaluated what happened according to a series of questions:

\begin{enumerate}
    \item Did the robot successfully move the plastic water bottle from the table into the trash? (Yes or No)\footnote{This question was shown for all videos. The others were only shown in the first, condition video.}
    \item Did the robot spend extended periods of time without making any progress?  (Yes or No)
\item Did the robot do something other than moving the water bottle into the trash? (Yes or No)
\item Did the robot have difficulty physically moving the water bottle? (Yes or No)
\item If the robot failed to complete the task, can you describe what happened? (Short Answer)\footnote{This question was only shown if the participant answered `No' to Q1.}

\end{enumerate}

The participant then repeats the same betting procedure with two success videos. After viewing all three videos, the user was asked to evaluate the odds of success of the robot in the future, along with a confidence score and an open question about their reasoning---similar to the betting screen, but without the actual bet. 

After every bet, participants were informed about the remaining payout, which decreased by \$2 every time they lost betting on the robot. 
Coin flips were counted throughout the study, but only performed once at the end of the study.

\subsection{Experimental conditions}
We examined 11 conditions split into three categories:  Failure, failure with subsequent success (\say{Failure+Success}), and Success. For failure conditions, we examine slip, lapse, and three different types of mistakes. \footnote{The original protocol included three failure conditions---a slip, a lapse, and a mistake in which the robot placed a thermos, instead of the water bottle, into the trash.
We found that 42\% of participants interpreted the thermos mistake as a successful execution. To better understand how users perceive an actual mistake condition, we added two more versions: placing a stapler in the trash and placing a marker in a cup.
This resulted in a total of five failure conditions and eleven total conditions.} For Failure+Success conditions we examine each of these failures immediately followed by a success. Finally, we examine direct success with no preceding failure. We assign participants evenly to all conditions through balanced randomization. Videos for each condition are attached as supplemental material. 

\subsubsection{Failure Conditions}
Participants in the Failure category saw one of the five failure conditions below, also summarized in Fig.~\ref{fig:conditions}.
\begin{enumerate}
    \item \textbf{Slip:} attempting the requested task, but failing to mechanically execute it. The robot fails to open its gripper, pushing the bottle instead of picking it up.
    \item \textbf{Lapse:} extended periods of time spent without making any progress on a task. The robot moves towards the bottle, then freezes, eventually returning to its resting position.
    \item \textbf{Mistake (Thermos):} completing a different task. The robot places a reusable thermos into the trash can instead of the water bottle.
    \item \textbf{Mistake (Stapler):} The robot places a stapler in the trash can instead of the water bottle. 
    \item \textbf{Mistake (Marker):} The robot places a dry erase marker into a mug instead of placing the the water bottle into the trash can.
\end{enumerate}



\subsection{Dependent Variable: From Bet to PR}
After each video, we are interested in measuring the robot's PR, which we define as the participant's belief that the robot is going to succeed in its task. 
To recap, we ask users to choose between a \say{fair coin} lottery, which has a 50\% chance of success, and betting on the robot. In each case, if the lottery loses or the robot fails, respectively, the participant loses \$2 out of their total \$10 study participation payout. 
Participants were also asked to state their confidence on a scale of 1 to 5  in the choice of their bet. 

The PR score used in the remainder of this paper is the participant's \say{signed confidence}~\cite{aitchison2015doubly}, calculated as the binary decision (-1 for lottery, 1 for robot) multiplied by the participant's confidence score. This is equivalent to an 11-point Likert item, excluding level 0, from \say{very confident in the lottery} ($-5$) to \say{very confident in the robot} ($5$). 

While there is significant debate in how to understand and measure a person's confidence in a decision or their belief in an outcome~\cite{kepecs2012computational, pouget2016confidence}, we use existing practices to interpret \say{signed confidence}~\cite{aitchison2015doubly} as the participant's confidence in the robot's success.
Choosing lotteries is an established mechanism in behavioral economics~\cite{holt2002risk}, which---under common assumptions of rationality---states that betting on the robot reflects at least a 50\% belief that the robot would succeed. 
The confidence measure then correlates with the distance from this 50\% mark. For more discussion on the signed confidence measure, see, e.g.,~\cite{mamassian2022modeling}.

In contrast to most HRI studies, which rely primarily on subjective measures without any real stakes for participant choices, our PR metric design uses a strictly objective measure, and one with monetary risk associated with it, similar to~\cite{kshirsagar2019monetary}.

Finally, we measure PR before and after seeing a video with the experimental condition to determine the impact of each condition. This also controls for dispositional difference in each participant by treating the first bet as a baseline.

\subsection{Statistical Testing}
We preregistered three different statistical tests on the difference between PR before and after the condition video: 1) a one-way ANOVA comparing all Failure conditions, 2) a one-way ANOVA comparing all Failure+Success and Success conditions, and 3) an independent Welch's t-test comparing conditions without a success (Failure) against the conditions with a success (Success and Failure+Success)\footnote{Code and output for all statistical tests can be found at www.github.com/violetteavi/impact-of-different-failures-analysis}. We meet normality assumptions by collecting 30 samples from each condition. Both success and failure ANOVA groups satisfy homoscedasticity assumptions (Levene's test, F(4,140)=0.97, $p=0.43$ for failure and F(5,175)=0.96, $p=0.44$ for success). To control for the rate of false positives, we compute $q$-values by applying Holm-Bonferroni \cite{holm_sture_simply_1979} correction to $p$-values of the three tests. Our threshold for strong evidence of an effect is $q<0.05$.

In addition to the preregistered tests, we ran an exploratory independent t-test on the final bet (after seeing all videos) to determine if failure has a lasting impact. This test compared the  conditions without a success (Failure) against the conditions with a success (Success and Failure+Success). This was not included in the false-positive correction calculation.

\subsection{Affinity Diagramming}
To better understand participant betting patterns, we used affinity diagramming \cite{abascal_using_2015} on the free response portion of each bet. We capture key observations using quotes from participants' responses,  then group quotes into clusters to extract larger trends. Finally, we convert the larger trends into codes, and quantify how many participants fit each trend. 

\subsection{Participants}

We recruited participants through the online recruitment platform Prolific\footnote{ www.prolific.com}. We collected data from 357 participants, of which 20 did not complete the survey and 11 were removed for suspected large language model (LLM) usage. Thus, we analyze the data of 326 participants (161 male, 159 female, 6 other, aged 18-78, median:37). Participants started with \$10 (USD) in their `compensation account'. Since participants placed three bets, each worth \$2, they received between \$4 and \$10 depending on their bets, with an average compensation of \$8.61.

To eliminate inattentive participants and clean up our qualitative response pool, we screened participants by detecting Generative AI usage based on four criteria: The first is including common parts of LLM output, such as \say{Great! Here's a sample explanation you can use or adapt [...]} The second is writing in the second person, such as \say{Opting for the 50/50 coin flip suggests that you preferred a guaranteed probability rather than relying on the robot’s past performance.}  The third is referencing videos that have not been shown. For example, one participant wrote this before seeing any videos: \say{According to the earlier videos, the robot seemed to be fairly steady in its efforts [...]} The last criteria is talking about robot demonstrations as if they have not happened. For example, one participant wrote this after seeing three robot demonstrations: \say{After observing the prior two setups and performance patterns (assuming some demonstration occurred), there may now be enough evidence to judge the robot’s reliability.} Generally speaking, LLM-generated responses seem to be written by a third party with only secondhand knowledge of the experiment.



\section{Findings}
\label{sec:Findings}

\begin{figure*}[t]
  \includegraphics[width=1\linewidth]{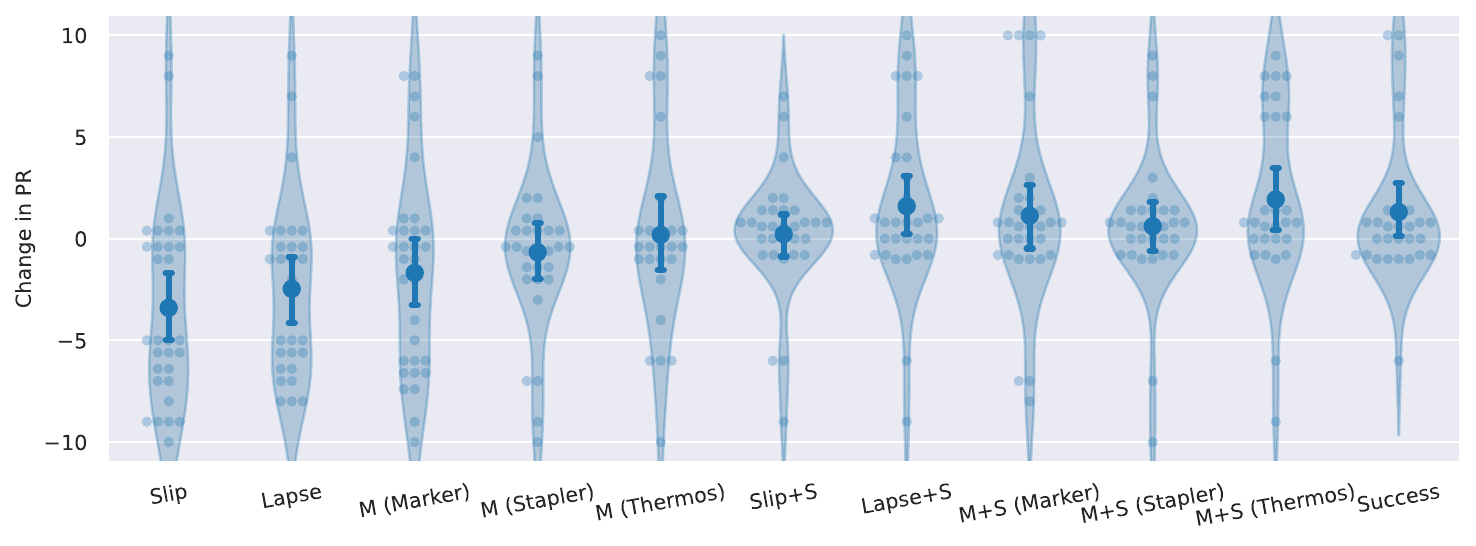}
  \caption{Change in participant bet after they see the condition video, compared to the bet before. 
  Each light dot represents one participant, the dark dot is the mean of participants, and the whiskers are the 95\% confidence interval of the mean. We see evidence for differences in effect of failure conditions (slip, mistake, and lapse), with no evidence for difference in effect if the demonstration contains success. Mistake and Success are abbreviated as M and S, respectively.
  }
  \label{fig:BetDiff}
  \vspace{-2mm}
\end{figure*}


Our main pre-registered hypotheses (Section~\ref{sec:Methods}) propose that different types of failures affect PR in distinct ways, and that successes will have different impacts on PR depending on whether a failure precedes the success and depending on its type.
In this section, we present quantitative findings testing these hypotheses using the PR measure described above, and contextualize these results with qualitative coding from participants’ responses to open-ended questions.
We then present findings from our surveys that offer additional insights on how participants perceived robot failures.

\subsection{Type of failure matters}

We find that different types of failure had different effects on PR. 
As preregistered before the study, we calculate $\Delta^{1\rightarrow2}_{\text{PR}}$, the difference in PR score between the second bet (after seeing the conditional video) and the first bet (which can be considered a participant's baseline).
This measure gives us the specific change in PR for each experimental condition.
Comparing this measure across the five failure conditions, we find weak evidence for a difference (One-Way ANOVA, F(4,140)=2.677, $p=0.034$, $q=0.068$). 
Fig.~\ref{fig:BetDiff} shows  $\Delta^{1\rightarrow2}_{\text{PR}}$ for all conditions; failures are on the left. 

\begin{table}[t]
\centering\rowcolors{2}{gray!25}{white}
\begin{tabular}{@{\extracolsep{0pt}}p{2.50cm}p{2.10cm}p{2.10cm}}
\toprule   
 & {Failure} & {Failure+Success}\\
\midrule
Slip & $-3.40\pm0.84$ & $+0.23\pm0.69$ \\
Lapse & $-2.46\pm0.87$ & $+1.60\pm0.70$ \\
Mistake (Marker) & $-1.68\pm0.82$ & $+1.13\pm0.69$ \\
Mistake (Stapler) & $-0.67\pm0.84$ & $+0.61\pm0.69$ \\
Mistake (Thermos) & $+0.19\pm0.90$ & $+1.93\pm0.71$ \\
No Failure &  & $+1.31\pm0.71$ \\
\bottomrule
\end{tabular}
\captionof{table}{ $\Delta^{1\rightarrow2}_{\text{PR}}$, the difference in perceived reliability (PR) after seeing the conditional video across conditions, reported as mean $\pm$ standard error.
}
\label{tab:bet21}
\vspace{-2mm}
\end{table}

Of the failure modes, Slip and Lapse caused the largest decreases in PR ($-3.40\pm0.84$ and $-2.46\pm0.87$, respectively). 
Mistake had the least detrimental effect on PR. 
In the Mistake (Thermos) condition, PR was nearly the same after seeing a failure ($+0.19\pm0.90$, see Fig.~\ref{fig:BetDiff} middle). 
Mistake (Stapler) and Mistake (Marker)  caused more modest decreases in PR  than Slip and Lapse ($-0.67\pm0.84$ and $-1.68\pm0.82$, respectively).

Open responses support these results.
Almost half of the participants in the Slip and Lapse condition cited past failure as a reason to bet against the robot (14/30 participants and 12/28, respectively). 
For example, P88 of the lapse condition said \say{The robot failed a task very similar to this one in the first trial.}
In contrast, much fewer participants in the Mistake (Thermos) 
and Mistake (Stapler) conditions wrote that failure influenced their bet (5/26 for thermos, 3/30 for stapler).

\subsection{Mistakes are not all the same}

It is worth noting the difference in $\Delta^{1\rightarrow2}_\text{PR}$  between the three types of mistakes.
Placing a reusable thermos in the trash resulted in no harm to PR ($+0.19\pm0.90$), and putting a stapler in the trash caused a negligible decrease in PR ($-0.67\pm0.84$). The biggest decrease in PR for a mistake occurred for placing a marker in the cup. We present possible explanations for this finding in the Discussion.

\subsection{Given eventual success, failures do not matter}

Comparing the combined average $\Delta^{1\rightarrow2}_{\text{PR}}$ for all Failure conditions (M=$-1.64$, SD=3.83) to the average for the combined data including all Failure+Success conditions and the direct Success condition (M=$+1.12$, SD=4.72), we find, as expected, that the latter is markedly better for PR than the former ($\delta=2.76\pm0.48$, t(324)=5.84, $p<0.001$, $q<0.001$). Contrast, for example, Fig.~\ref{fig:BetDiff} left and right.
Any success in the video increases the average PR compared to the baseline.

This is also supported by the open response coding---participants cite the robot's eventual performance when justifying their bets:
43\% of participants in the Failure+Success cases (78/181) indicated that prior success was the primary reason they bet on the robot. For example, P42 (Lapse+Success) said \say{After seeing the previous robot being able to remove a water bottle [...] I have faith in the robot's success.} In contrast, 37\% of participants in the Failure conditions (54/145) indicated that prior failure was the primary reason they bet against the robot---P88 (Slip) wrote that \say{The robot failed earlier, let me see whether the coin flip will at least be a better bet.}

If the robot succeeds, the different conditions are statistically indistinguishable. Comparing $\Delta^{1\rightarrow 2}_{\text{PR}}$ for all Failure+Success conditions along with the direct Success condition, did not suggest any difference (One-Way ANOVA, F(5,175)=0.81, $p=0.543$, $q>1$). 

\subsection{Success is an effective recovery method}

\begin{figure}[t]
  \centering
  \includegraphics[width=1\linewidth]{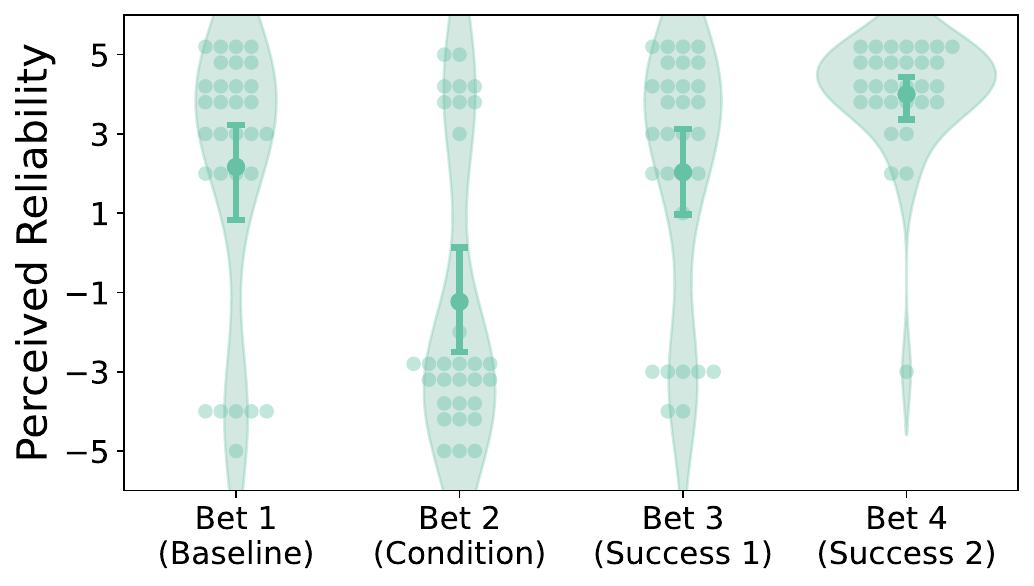}
  \caption{Participant betting behavior before any videos (bet 1), after the Slip video (bet 2), and after additional success videos (bets 3 and 4). Perceived reliability (PR) drops after seeing the robot slip. PR rises after seeing success, ending above the baseline from Bet 1.} 
  \label{fig:Recovery}
\end{figure}

We find that seeing successful execution after the conditional video is an effective recovery technique, matching existing literature~\cite{pan_how_2023}. 
This trend can be seen in Fig.~\ref{fig:Recovery}, even for the biggest drop in PR (Slip). 
Nearly all participants (310/326, 95\%) bet on the robot with a confidence rating of at least 3/5 after two success videos. 

This trend is also reflected in the free response data for the failure cases. 
Almost no participants in Failure conditions (5/145, 3\%) indicated, directly after seeing the failure, past success as the reason they would bet on the robot. 
This rose to 62\% of participants in Failure conditions (90/145) after seeing two successes. 
P342 (Mistake (Marker)) illustrates this: \say{The robot has demonstrated twice after the initial failure that it is capable of performing the task. Thus, the error from the first attempt has been resolved.}
All conditions showed a higher PR than the participants' baseline after two successes. 
However, there remained a gap in PR between the Failure (M=3.88, SD=1.46) and Failure+Success/Success (M= 4.58, SD=0.87) conditions after two successful robot executions ($\delta=0.70\pm0.13$, $t(324)=5.34$, $p<0.001$).


\subsection{Failure Perception}
\label{subsec:perception}

\begin{figure}[t]
  \centering
  \includegraphics[width=0.86\linewidth]{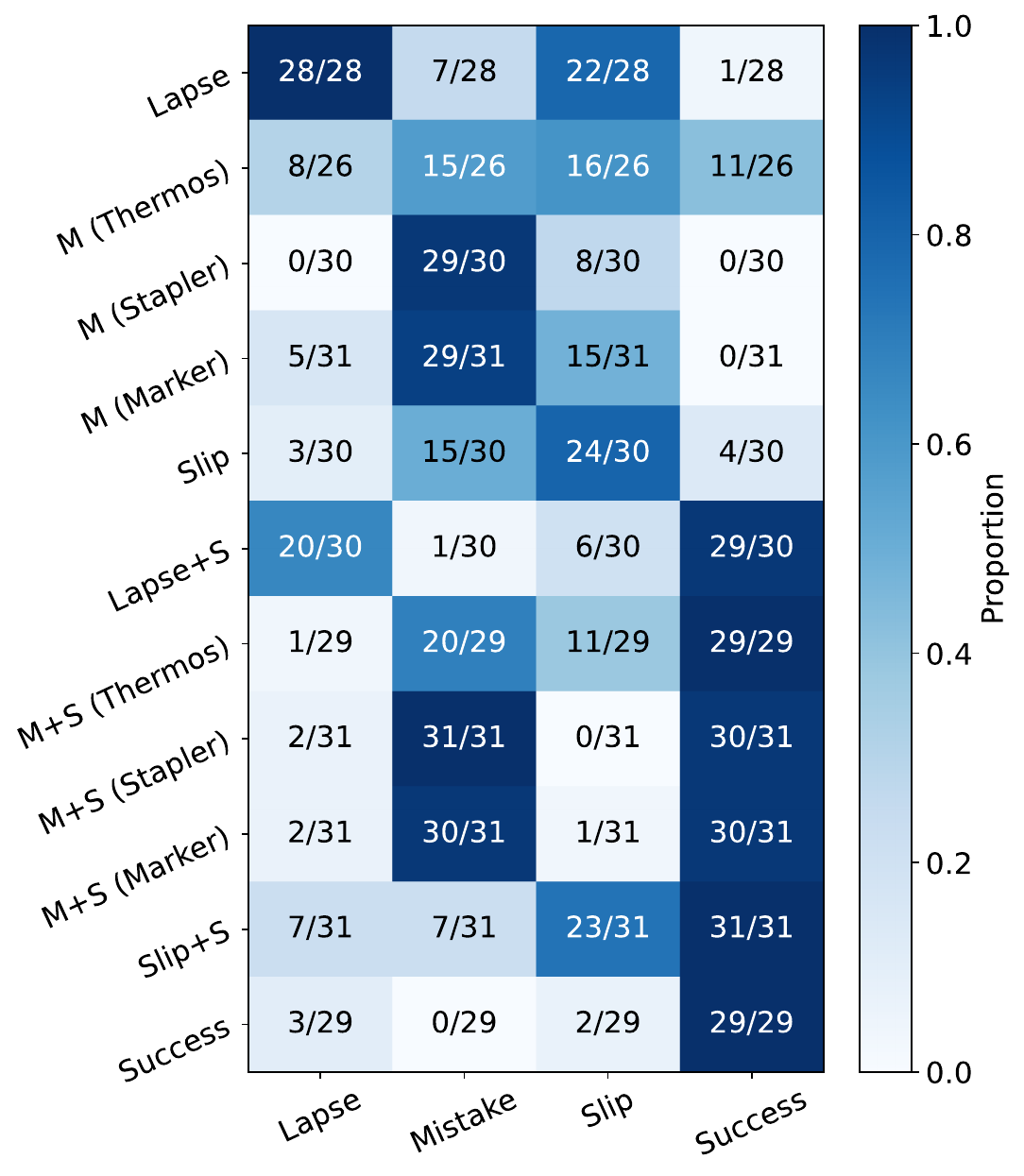}
  \caption{Participant evaluation (horizontal) matched our labels (vertical), though participants often labeled lapses and mistakes as slips. Mistake (M) and Success (S) are abbreviated.}
  \label{fig:Confusion}
\end{figure}

Finally, we analyze the responses to the survey questions in which participants were asked to describe what happened in the video before learning the outcome of their bet.
To determine how participants perceive the robot's performance, we compare the experimental conditions as we intended them to a series of questions answered by participants (see Section~\ref{subsec:Flow}). Participants tended to agree with the conditions' labels of failure modes, with some notable exceptions. Fig.~\ref{fig:Confusion} shows a breakdown of these disagreements.

Participants agreed on the success label, with 94\% (307 of 326) participants matching the experimenter's labels of the video. 
The `success' label disagreement largely lies in the Mistake (Thermos) condition, in which the robot places a reusable thermos with a handle into the trash. 42\% of participants viewed this as a successful execution, despite the disposable plastic bottle remaining on the table.

Another area in which our label did not agree with participants' interpretations is that many applied the statement that the robot had \say{difficulty physically moving the water bottle}---which we associated with the Slip condition---to Mistakes and Lapses. 
For 29\%  of participants (95 out of 326), there is a mismatch between their condition and their response to the above statement.
This mismatch primarily comes from participants labeling other failure modes as slips. 
This number is lower for Lapse (13\%) and Mistakes (17\%).

\section{Discussion}
\label{sec:Discussion}

Why do the different types of mistakes lead to different drops in PR?  While slips and lapses notably decrease PR, mistakes are overall less harmful to PR, but some mistakes are worse than others. 
We have limited information, having compared only three types of mistakes, but can speculate possible explanations for the difference between mistakes. 

First, the severity of loss in PR  may be related to the semantic similarity of the mistaken task relative to the intended task. Participants may view putting a thermos into the trash more positively than putting a stapler into the trash because the thermos, like a bottle, is an object that can contain liquids for drinking.

Alternatively, a mistake in object handled might be less crucial than one in the goal position. So, as long as the goal is correct, even if the wrong object was picked up, the overall task seems doable by the robot. 
This could explain why putting the wrong object in the trash indicates higher reliability than putting the wrong object in a mug. 

A third explanation is an implicit breakdown of the task into smaller pieces---the participant may separate pick-and-place into a pick, and a place. In the Mistake (Thermos) and Mistake (Stapler) conditions, the robot may get partial credit for the \say{place} part. The Mistake (Marker) condition would not get this credit, as it picks the wrong object and places into the wrong location.

Open responses suggest that mistakes can indeed demonstrate some capability, even if they do not achieve the task.
For example, P308 of the Mistake (Stapler) condition wrote: \say{I saw it successfully move the stapler in the trash, so I believe in its abilities but not so sure it will pick up the right item... I know it is capable but not sure overall.}

Some participants indicated that the Mistake (Marker) case occurred due to disobedience instead of a lack of capability. This raises questions around what is considered a robot's motivation versus its ability, a known distinction in human trust theory \cite{malle_multidimensional_2021}. When thinking about a robot's ability, do humans assume that correctly processing language instruction is an inherent robot capability, perhaps due to the proliferation of strong natural language systems? 
Does this expectation extend to other semantic tasks, such as object identification via computer vision? 

Finally, the mismatch between condition labels and user interpretations (Section~\ref{subsec:perception}) suggest that people consider \say{cognitive} failures, such as lapses or mistakes as \say{physical} breakdowns. 
This might indicate, again, that people working with robots may distinguish between what a robot knows and what a robot knows how to do.

In summary, how do failures matter? Though the initial effect of each failure mode was different, successful executions recovered PR.  This provides evidence for leniency:
As long as the task is eventually completed,  participants were accepting of failures that occurred along the way. 
However, this recovery does not imply robustness to future failure. Recency bias in HRI is well-documented \cite{luebbers_recency_2024}. 
PR is unlikely to remain high if the robot fails repeatedly. Understanding which failures need mitigation is still important. 

\subsection{Limitations and Future Work}
Our study has several limitations.
First, participants get paid based on whether the study labels the robot behavior as a success, and they see this judgment in real time. 
Therefore, the betting measure might not necessarily measure the user's perception of robot probability to succeed according to their own standards, but their desire to maximize compensation by guessing what the experiment considers a success. That said, only 19 out of 326 participants (6\%) labeled success differently from the experiment design, so it is unlikely that these differences significantly impacted our findings.

Another limitation is that our study was conducted online, which may not generalize to in-person HRI. Some experiments show that effects in online studies are mirrored in physically situated studies, and only vary in the magnitude of the effect (e.g.,~\cite{hu2019using}). Others suggest more qualitative differences between video and in-person interactions with robot (e.g.,~\cite{bretan2015emotionally}). In both cases, there may have been effects we were unable to detect.


Finally, while this study illuminates differences in PR after failure, it does not provide participants with the ability to interact with the robot. Interactivity may change the results of the study either through premature termination of the robot's activity or even just through a stronger sense of control over the robot. 
For example, if the human sees the robot grabbing a stapler instead of the water bottle, they may stop the execution. This would convert a Mistake+Success into a Mistake, as the robot does not have the opportunity to succeed once a failure occurs. The timing of human intervention during these failures could be further studied.


\section{Conclusion}
\label{sec:Conclusion}

In this study, we compared different robot failure conditions using a pre-registered and controlled online video experiment. Our findings reveal that not all robot failures are perceived equally when it comes to the robot's perceived reliability (PR): physical slips are most detrimental, followed by robot freezes (lapses). The failures that were least harmful for the robot's PR were mistakes that could be attributed to object recognition or plan errors. 

All that said, we also find that, when failures are followed by a successful execution of the task, they do not harm PR and are statistically indistinguishable from successes. Moreover, PR recovers rapidly when the robot succeeds after a failure, even in the absence of explicit repair behaviors, such as apologies or explanations.

This result highlights a critical nuance in human-robot interaction: not all robot failures are equal and need to be treated with the same repair mechanism. Recovery from failure does not even necessarily require overt repair mechanisms. Instead, consistent successful performance can restore a robot's reliability perceptions over time. These insights shed light on which types of robot failures matter most in shaping user perceptions and inform priorities when designing interaction strategies that mitigate the impact of failure.

Moreover, our findings motivate existing questions about how people interpret robot failures—--particularly in terms of perceived robot ability versus motivation, and the distinctions between physical and cognitive capabilities. Understanding these interpretive frames is essential for developing robots that can effectively manage user expectations and maintain trust in human-robot interaction.










\bibliographystyle{IEEEtran}
\bibliography{references}

\end{document}